\documentclass[conference]{IEEEtran}
\IEEEoverridecommandlockouts
\usepackage{cite}
\usepackage{amsmath,amssymb,amsfonts}
\usepackage{algpseudocode}
\usepackage{algorithm}
\usepackage{graphicx}
\usepackage{hyperref} 
    
\begin{document}
\title{A Simple and Effective Temporal Grounding Pipeline for Basketball Broadcast Footage\\}
\author{\IEEEauthorblockN{Levi Harris}
\IEEEauthorblockA{\textit{Department of Computer Science} \\
\textit{The University of North Carolina at Chapel Hill}\\
Chapel Hill, USA\\
levlevi@cs.unc.edu}
}

\maketitle{}
\begin{abstract}
We present a reliable temporal grounding pipeline for video-to-analytic alignment of basketball broadcast footage. Given a series of frames as input, our method quickly and accurately extracts time-remaining and quarter values from basketball broadcast scenes. Our work intends to expedite the development of large, multi-modal video datasets to train data-hungry video models in the sports action recognition domain. Our method aligns a pre-labeled corpus of play-by-play annotations containing dense event annotations to video frames – enabling quick retrieval of labeled video segments. Unlike previous methods – we forgo the need to localize game clocks – fine-tuning an out-of-the-box object detector to find semantic text regions directly. Our end-to-end approach improves the generality of our work. Additionally, interpolation and parallelization techniques prepare our pipeline for deployment in a large computing cluster. All code is made publicly available:
\url{https://github.com/leharris3/contextualized-shot-quality-estimation/tree/temporal-grounding-pipeline}

\end{abstract}

\begin{IEEEkeywords}
objection detection, objection recognition, optical character recognition
\end{IEEEkeywords}

\section{Introduction}
\subsection{Motivation}
Text-to-video temporal grounding is a growing field of interest and essential to developing large, multi-modal video datasets \cite{b1}, \cite{b2}. In particular, fine-grained temporal alignment of sports broadcast footage and play-by-play annotations spans academic and industrial interests. Sports action recognition benchmarks offer a popular method of evaluating video model capabilities in academia. Multi-sport and sport-specific datasets \cite{b3} provide practical test beds for state-of-the-art video understanding systems. In particular, we are interested in basketball broadcast video datasets aligned to play-by-play annotations (see Figure 1). These datasets offer enormous cost savings and potential for scalability – due to the ready availability of play-by-play annotations and broadcast clips through content providers like Hudl. Mainly, deeply indexed game footage enables rapid, semi-supervised generation of action-recognition datasets in a semi-supervised manner.

While not the focal point of our work, our method may also be applicable in an industrial setting. Temporal alignment of game broadcasts and play-by-play logs may aid the development of automatic highlight detection systems, player recognition/tracking software \cite{b4}, automated game video annotation pipelines, and segmentation and analysis tools.

\begin{figure}[h]
\centering
\includegraphics[width=\linewidth]{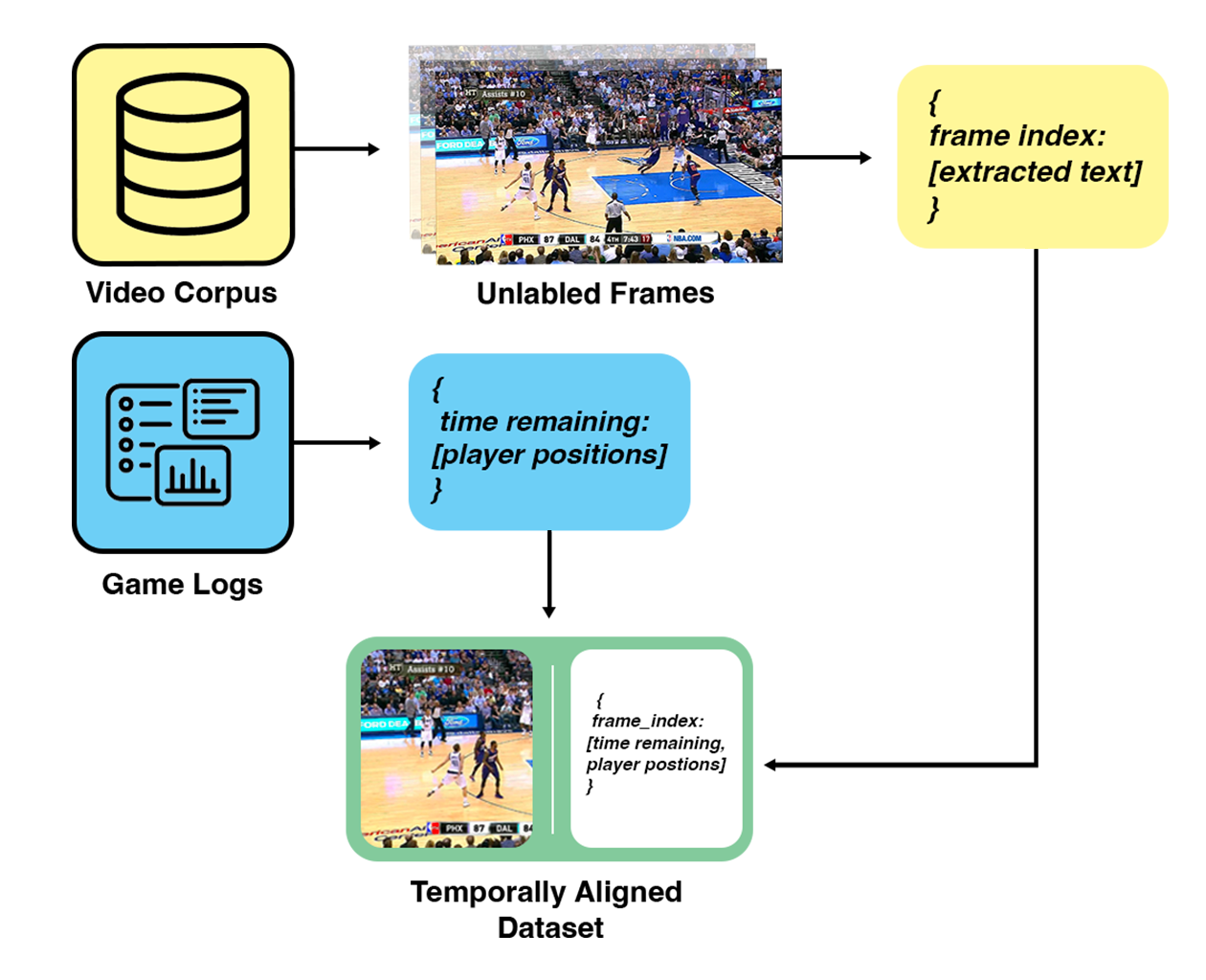}
\caption{We motivate our pipeline with the figure above. Given pre-annotated game logs and unlabeled video footage, we devise a method to align both modalities. The final result of our temporal grounding pipeline is a video-to-text aligned corpus intended to enable rapid dataset development.}
\label{fig:figureone}
\end{figure}

\subsection{Challenges}
Extracting text from broadcast scenes poses several unique challenges. Primarily, ensuring the generality of a text-extraction method for many different broadcast providers, basketball leagues, and video resolutions is especially difficult. Basketball broadcasters use unique game clock graphics and fonts – making any hard-coded solution impractical. Additionally, our unprocessed video corpus includes irrelevant moments (ad breaks, breaks in play) and unpredictable occlusions of key text regions. We meet these challenges by training our custom text detection model on a robust, proprietary dataset. Additionally, we rigorously test our pipeline in a custom test bed before deployment.

\section{Related Work}

Previous work in text extraction from sports broadcast footage spans many use cases and methodologies. The authors of \cite{b5} use Optical Character Recognition (OCR) to extract quarter and time remaining values from basketball broadcast scenes – within the context of a player detection and tracking system. While the authors claim to achieve high accuracy by their methods – they hard-code text region positions. In our work, we use object detection methods to find our Regions of Interest (ROI). In \cite{b6}, the authors use a two-staged text-extraction approach for the temporal alignment of sports broadcast footage and play-by-play game logs. First, for each frame in a video game, clock regions are detected using a single-shot object detector with a CGG16 backbone trained on a custom dataset of game clocks. Next, the authors perform text recognition using an in-domain trained CRNN to extract unlabeled strings. Finally, in-domain Knowledge Constraints (KC) ensure that extracted text conforms to sport-specific regular expressions. For example, a quarter string must conform to a regular form (e.g., 3rd.) KCs allow the authors to match unlabeled text to semantic text labels. Their method, while accurate, requires an additional step to map unlabeled text strings to semantic text labels. We successfully remove this step by directly finding semantic text regions using an in-domain trained object detector. \cite{b7} describes a large dataset of soccer broadcast videos with temporally labeled events. The others design a text-extraction pipeline to align play-by-play event labels to unlabeled game footage. The authors conduct a statistical study of pixel densities in a video to identify candidate regions for the game clock. OCR is then used to extract text strings from the ROIs. Finally, a RANSAC approach ensures the temporal consistency of extracted text values. The presented text recognition method is unlikely to generalize beyond soccer broadcast footage due to its reliance on hand-crafted features. 

Of the text-extraction methods previously mentioned, \cite{b6} provides the best approach. The outlined approach generalizes well to many different sports-broadcast types –  and provides the primary inspiration for our work. In contrast, our text-extraction pipeline detects semantic text regions directly. Moreover, the authors conduct their research in an industrial setting – and thus do not provide a detailed implementation of their method. After reviewing the relevant literature, we conclude our text-extraction method for basketball broadcast scenes is superior to existing methods in simplicity of design and ease of implementation.

\section{Approach}
\subsection{Curating a Custom Text ROI Dataset}

\begin{figure}[h]
\centering
    \begin{algorithm}[H]
    \caption{Extract Timestamps from Broadcast}
    \begin{algorithmic}[1]
    \State \textbf{Input:} video path
    \State \textbf{Output:} timestamps
    \State timestamps $\gets$ \text{empty list}
    \State \textbf{procedure} \text{LOAD VIDEO}(video path)
    \State \hspace{\algorithmicindent} \text{Open video file at} video path
    \State \textbf{end procedure}
    \State \textbf{procedure} \text{PROCESS VIDEO}()
    \For{each frame in video}
        \State \textbf{ROI Detection:}
        \State \hspace{\algorithmicindent} roi frame $\gets$ \text{extract ROI from frame}
        \State \textbf{Text Extraction:}
        \State \hspace{\algorithmicindent} timestamp $\gets$ \text{extract text from} roi frame
        \State \hspace{\algorithmicindent} \text{append} timestamp \text{to} timestamps
    \EndFor
    \State \textbf{end procedure}
    \State \textbf{procedure} \text{FINALIZE}()
    \State \hspace{\algorithmicindent} \text{Close video file}
    \State \hspace{\algorithmicindent} \textbf{return} timestamps
    \State \textbf{end procedure}
    \end{algorithmic}
\end{algorithm}

\caption{Algorithm demonstrating our two-staged approach to text-extraction from basketball broadcast scenes. We first load a video into a buffer object. Next, for each frame in a video, we recognize semantic text regions using an in-domain text classifier. Second, we pass cropped text regions to our out-of-the-box text recognition model. We omit details of our parallelization scheme and post-processing algorithm for brevity. Full implementations of our algorithms are available in our repository.}
\end{figure}

Our text-extraction pipeline adopts a two-staged approach – text detection with a custom object-detection model and text recognition with an out-of-the-box OCR tool. Given a series of video frames as an input, we directly localized semantic text ROIs using a fine-tuned YOLOv8l object detection model [yolo paper]. Before training, we collected approximately 30,000 sample frames from basketball broadcast footage from Hudl. We sample games from the NBA, WNBA, Euroleague, NCAA, WNCAA, and various American high school basketball leagues. Next, using the CVAT annotation tool [informal citation of CVAT], we match each video frame with a quarter and time-remaining object bounding box label. Using the static properties of broadcast clocks, we reduce the time required to annotate our dataset. CVAT allows users to quickly annotate videos by defining an object region in a frame and then playing a video to annotate bounding boxes. We fine-tune a pre-trained model for 130 epochs with all training parameters set to their default values by the Ultralytics library. We employ heavy image augmentation to avoid overfitting and improve model generality. Training took approximately 8 hours on a single T4 GPU.

\subsection{Text ROI Detection with YOLO}

We read in frames from a video buffer at a fixed step interval, denoted as \( t \). Next, we localize semantic text regions by predicting bounding boxes with our YOLO model. We fix a confidence threshold: \( T \). Detection halts if confidence scores \[ C = Pr(\text{object}) \times IoU \] exceed T for both text regions for a given frame. Our algorithm returns an empty dictionary if no valid text regions are detected. Note that in most cases, we assume text ROIs to be static. We observed in the rare instances an ROI briefly changes positions or experiences an occlusion – we can interpolate between missing values during our post-processing stage.

\subsection{Text Recognition with PaddleOCR}

We use the PaddleOCR [informal citation] library out-of-the-box for text recognition. Given quarter and time-remaining ROIs, we crop each frame into a respective sub-region. Next, we pass each cropped text region to a PaddleOCR model and perform text recognition. We manually turn off the PaddleOCR text-detection feature, as we already perform this step in 3.1.2. PaddleOCR performs far better in practice for digital text recognition than other off-the-shelf toolkits such as PyTesseract and EasyOCR. After experimenting with several image pre-processing configurations – we discovered that PaddleOCR is most accurate on cropped frames resized to 90 DPI with no additional image pre-processing. Our pipeline extracts text with perfect accuracy from \( \frac{91}{97} \) (or about 93.81\%) frames randomly sampled from our dataset of basketball broadcasts before any post-processing is applied. 

\subsection{Denoising and Timestamp Interpolation}

We adopt our text-extraction pipeline for a corpus of basketball broadcast footage, pre-segmented by game period. For a given video, denoted as \( V \), we expect 'time remaining' values (\( T \)) to follow a negative linear trend during a game. Given that videos in our corpus maintain a consistent frame rate of 30 frames per second (fps), we can model the temporal progression as:

\[ T = T_0 - \frac{1}{30} \times n \]

where. \( T_0 \) represents the initial time remaining at the start of the period, and \( n \) denotes the number of frames elapsed.

We implement a two-stage denoising algorithm. First, we remove outlier values (those that lack temporal consistency) from our extracted dataset. This can be formulated as:

\[ T' = \{ t \in T | |t - \hat{t}| < \theta \} \]

where \( T' \) is the denoised set of time values, \( \hat{t} \) is the temporally consistent estimate of time at any point, and \( \theta \) is a threshold for determining outliers.

Second, we interpolate between monotonically decreasing 'time remaining' values, which can be expressed as:

\[ T_{interp} = \text{interp}(T', x) \]

where \( T_{interp} \) represents the interpolated time values, and \( \text{interp} \) is the interpolation function applied over the dataset \( T' \) at points \( x \).

We provide an implementation of our algorithm in our original code base.

\subsection{Parallelization for Large-Scale Deployment}

We implement a parallelization scheme in our pipeline to distribute computation across multiple CPU threads. This simple addition of functionality yields substantial improvements to overall runtime performance. On a standard Macbook M3 processor, runtime performance downscales linearly for each additional thread of execution. For example, parallelization to 2 workers yields a 50\% reduction in runtime, 4 workers generate a 75\% reduction, and so on. Distribution across multiple computing devices (rather than threads) may unlock further improvements in runtime performance. We will provide incremental performance updates to the codebase.

\section*{Results and Runtime Performance}

Our pipeline passes visual inspection – successfully aligning game footage to 2D player positions in most cases. We anecdotally note that our system performs best in high-resolution, homogenous broadcast scenes. We encounter many examples of noisy, occluded, or missing game clocks due to the diversity and scale of our video corpus –  While we handle occlusions smoothly during our post-processing phase – we are forced to discard videos with instances of missing game clocks. Performing the temporal grounding of broadcast footage without on-screen graphical indicators is very challenging. Therefore, we leave this problem to future work.

\section*{ Conclusions and Future Work}

We present a simple and reliable temporal grounding pipeline for basketball broadcast scenes. Our use of end-to-end text localization methods offers practical improvements over previous methods of sports-broadcast text extraction—moreover, by using readily available, out-of-the-box text-recognition tools – we ensure the reproducibility of our work. We believe our pipeline is adaptable in many use cases and sub-domains. Specifically, we foresee that our work will allow other researchers to quickly and easily develop large, multi-modal datasets for action recognition models.

While the foundation of our approach is sound, much work remains to scale our pipeline in an industrial setting. Thorough tests and quantitative benchmarks will be necessary to evaluate the robustness of our methods. Additionally, custom, in-domain trained text recognition models can enhance future versions of this pipeline. We reiterate that this work does not represent an advancement to the state of the art. Instead, we showcase the power and simplicity of open-sourced libraries and deep learning methods in a specific problem domain. Contributions to our repository are greatly appreciated; we will make incremental improvements to our codebase over time.


\begin{thebibliography}{00}
\bibitem{b1} Hao Zhang, Aixin Sun, Wei Jing, and Joey Tianyi Zhou. "Temporal Sentence Grounding in Videos: A Survey and Future Directions. (2022)" arXiv preprint arXiv:2201.08071 
\bibitem{b2} Jonghwan Mun, Minsu Cho, and Bohyung Han. "Local-Global Video-Text Interactions for Temporal Grounding.(2020)" arXiv preprint arXiv:2004.07514 
\bibitem{b3} Wu, F., Wang, Q., Bian, J., Xiong, H., Ding, N., Lu, F., Cheng, J., \& Dou, D. (2022). "A Survey on Video Action Recognition in Sports: Datasets, Methods and Applications." arXiv preprint arXiv:2206.01038
\bibitem{b4} Mackowiak, R., Cheshire, P., Ming, L., Mazzeo, P. L., Direkoglu, C., Naushad Ali, M., Rao, B., Markoski, B., Zhu, X., Chengjun, L., GerkeKarsten, S., Schäfer, R., Li, B., Liu, H., \& Bhanu, B. "Optical Tracking in Team Sports. (2022)" ar5iv.org.
\bibitem{b5} Lu, W.-L., Ting, J.-A., Little, J. J., \& Murphy, K. P. (2013). Learning to Track and Identify Players from Broadcast Sports Videos. IEEE Transactions on Pattern Analysis and Machine Intelligence, 35(7), 1704–1716. https://doi.org/10.1109/TPAMI.2012.242
\bibitem{b6} Shah, A., Biswas, T., Ramadoss, S., \& Shah, D. S. (2021). Distantly Supervised Semantic Text Detection and Recognition for Broadcast Sports Videos Understanding. arXiv. Retrieved from https://arxiv.org/pdf/2111.00629.pdf
\bibitem{b7}Giancola, S., Amine, M., Dghaily, T., \& Ghanem, B. (2018). SoccerNet: A Scalable Dataset for Action Spotting in Soccer Videos. arXiv:1804.04527.
1989.
\end{thebibliography}
\end{document}